\documentclass[letterpaper,twocolumn,10pt]{article}
\usepackage{usenix-2020-09}

\usepackage{tikz}
\usepackage{amsmath}
\usepackage{algorithm}
\usepackage{algpseudocode}
\usepackage{filecontents}
\usepackage{xspace} 
\usepackage{amssymb}
\usepackage{booktabs}

\newcommand{\sys}{Lyanna\xspace}

\begin{document}

\date{}

\title{\Large \bf Make Every Draft Count: Hidden State based Speculative Decoding }

\author{
{\rm Yuetao Chen}\textsuperscript{\rm 1}\quad
{\rm Xuliang Wang}\textsuperscript{\rm 2}\quad
{\rm Xinzhou Zheng}\textsuperscript{\rm 3}\quad
{\rm Ming Li}\textsuperscript{\rm 2}\quad
{\rm Peng Wang}\textsuperscript{\rm 4}\quad
{\rm Hong Xu}\textsuperscript{\rm 1}
\\
\textsuperscript{\rm 1}The Chinese University of Hong Kong\quad
\textsuperscript{\rm 2}University of Waterloo\\
\textsuperscript{\rm 3}University of Science and Technology of China\quad
\textsuperscript{\rm 4}Unaffiliated
}
\maketitle

\begin{abstract}
Speculative decoding has emerged as a pivotal technique to accelerate LLM inference by employing a lightweight draft model to generate candidate tokens that are subsequently verified by the target model in parallel. However, while this paradigm successfully increases the arithmetic intensity of memory-bound inference, it causes significant compute inefficiency: the majority of draft tokens fail verification and are discarded, resulting in waste of computation.  Motivated by the goal of recollecting this wasted computation, we propose a novel system that transforms discarded drafts into reusable tokens. Our key insight is to perform auto-regressive prediction at the hidden states level and postpone the integrating token information after the hidden states generation, so the draft hidden states are not contaminated by incorrect tokens, enabling hidden state reuse. To implement such a system, first we introduce a draft model architecture based on auto-regressive hidden states, which preserves richer semantics than token-based drafters to facilitate draft repurposing. Second, we design an efficient token information injection mechanism that leverages our specialized draft model to construct high-quality draft token trees and enables resampling tokens from verification failures. Third, we eliminate the overhead hidden in our design to further maximize hardware utilization. We conducted extensive evaluations against various baselines, demonstrating up to a 3.3$\times$ speedup against standard speculative decoding.
\end{abstract}

\section{Introduction}

Large language models (LLMs) have demonstrated remarkable capabilities and become the core of various next-generation applications such as machine translation~\cite{he2023exploringhumanliketranslationstrategy, xu2024paradigmshiftmachinetranslation}, conversational chatbots~\cite{openai2024gpt4technicalreport,geminiteam2025geminifamilyhighlycapable}, code assistants~\cite{hui2024qwen25codertechnicalreport,deepseekai2024deepseekcoderv2breakingbarrierclosedsource} and autonomous agents~\cite{wu2023planeliminatetrack,bran2023chemcrowaugmentinglargelanguagemodels}. To generate the new token, an LLM performs a full forward pass that attends over all previously generated tokens across all transformer layers. To avoid recomputing attention states, serving systems maintain a key--value (KV) cache in GPU memory. Each decoding step repeatedly reads the LLM weights and an ever-expanding KV cache from high-bandwidth memory (HBM), while performing relatively limited arithmetic per byte, resulting in the memory-bound nature of auto-regressive decoding.

Given that each LLM decoding step is memory intensive, a natural optimization goal is to reduce the number of full forward passes of the large model per generated token. Speculative decoding has been proposed to target this goal. Instead of running the large \textit{target model}, i.e., the original LLM, once per token, a small \textit{draft model} first guesses several future tokens; the system then uses the target model to verify the entire block of proposed tokens in parallel, requiring only one forward pass. When the proposals are accepted, speculative decoding effectively amortizes the cost of the target model's forward passes over multiple tokens, thereby mitigating the memory-bound bottleneck of LLM inference.

The efficiency of speculative decoding critically depends on how many proposed tokens are actually accepted by the target model at each step, i.e., the \emph{acceptance length}. Intuitively, if the target model frequently agrees with the draft model for long stretches of tokens, the cost of each expensive target-model forward pass can be amortized over many generated tokens, yielding substantial speedups. To improve the acceptance length, recent work has proposed \emph{tree-based speculative decoding} schemes~\cite{10.1145/3620666.3651335}. Instead of producing a single chain of tokens, the draft model constructs a tree of candidate tokens that represents multiple possible paths of drafting. The target model then verifies the tree in parallel and is more likely to accept with the longer acceptance length.

Nevertheless, tree-based speculative decoding methods incur considerable waste. By design, they generate many draft tokens, exploring multiple branches of draft token trees that the target model may or may not accept. After verification, only the tokens that lie on the selected path are kept; all other tokens on rejected branches are discarded. As the branching factor or tree depth increases, the number of discarded draft tokens grows quickly. As a result, while tree-based speculative decoding can increase the probability of obtaining a long accepted continuation, it does so by aggressively generating, verifying and then throwing away a large number of draft tokens, along with the computation and memory bandwidth spent to produce them.

The goal of this work is to reduce computation waste in the drafting process. We achieve this by introducing a novel mechanism which is capable of reusing the hidden states generated by the draft model even after the verification failures. We observe that in existing systems wrong tokens are non‑reusable, and so are the hidden states that produced them, because those hidden states were themselves computed from incorrect tokens. Our key insight is that, to reuse hidden states, the hidden states should be independently generated from tokens and the information of tokens should be leveraged after the generation of hidden states. Based on this, we present a novel speculative decoding system \sys, a hidden states based speculative decoding. However, there are several considerable challenges that must be tackled.

Firstly, the native hidden-state-based draft model which does not take the tokens as inputs loses the token information, which decreases the acceptance length and the performance of speculative decoding. Secondly, following the key insight outlined above, we defer token information integration until the sampling phase. The second challenge then becomes how to integrate token information that enables rejected drafts reuse in a manner that is training-efficient, and barely adds overhead to the sampling process. Moreover, a third challenge emerges as the algorithm's implementation may introduce performance-affecting conflicts at both the algorithmic and system levels. These include the conflict between verifying reused hidden states versus performing standard draft verification, as well as the trade-off between memory footprint and computation overhead.

\sys consists of three key techniques to address the aforementioned challenges: (1)~\textit{Draft Model Adaptation} for accurate prediction. We propose next hidden state prediction initialized with target model provided context and enhanced by posterior token context injection to maximize acceptance length. (2)~\textit{Token-info Injected Sampling} for efficient token information integration. We introduce token-info embeddings which integrates token information without incurring additional overhead in the first sampling and the subsequent re-sampling process to construct draft trees from the hidden state chain. (3)~\textit{Overhead Removal} for performance optimization.  We break the trade-off between memory footprint and computation by exploiting hot-token sparsity and fusing verifications to further improve performance.

We have conducted extensive evaluations to compare
\sys with other state-of-the-art speculative decoding algorithms integrated into the popular LLM serving systems. The results
show that \sys consistently outperforms all baselines.
Specifically, \sys achieves up to a 3.3× throughput compared to the standard speculative decoding and a 1.4× throughput compared to the best-performing baseline. Detailed ablation experiments in both model and system aspects demonstrate the effectiveness of our designs. Finally, based on the breakdown experiments, \sys reduces the draft model's forward pass latency by 60.9\%.

\section{Background and Motivation}

We start by presenting the brief background of speculative decoding, followed by the motivation of recycling the rejected drafts. 

\begin{figure*}[thp]
\begin{center}
\includegraphics[width=0.95\linewidth ]{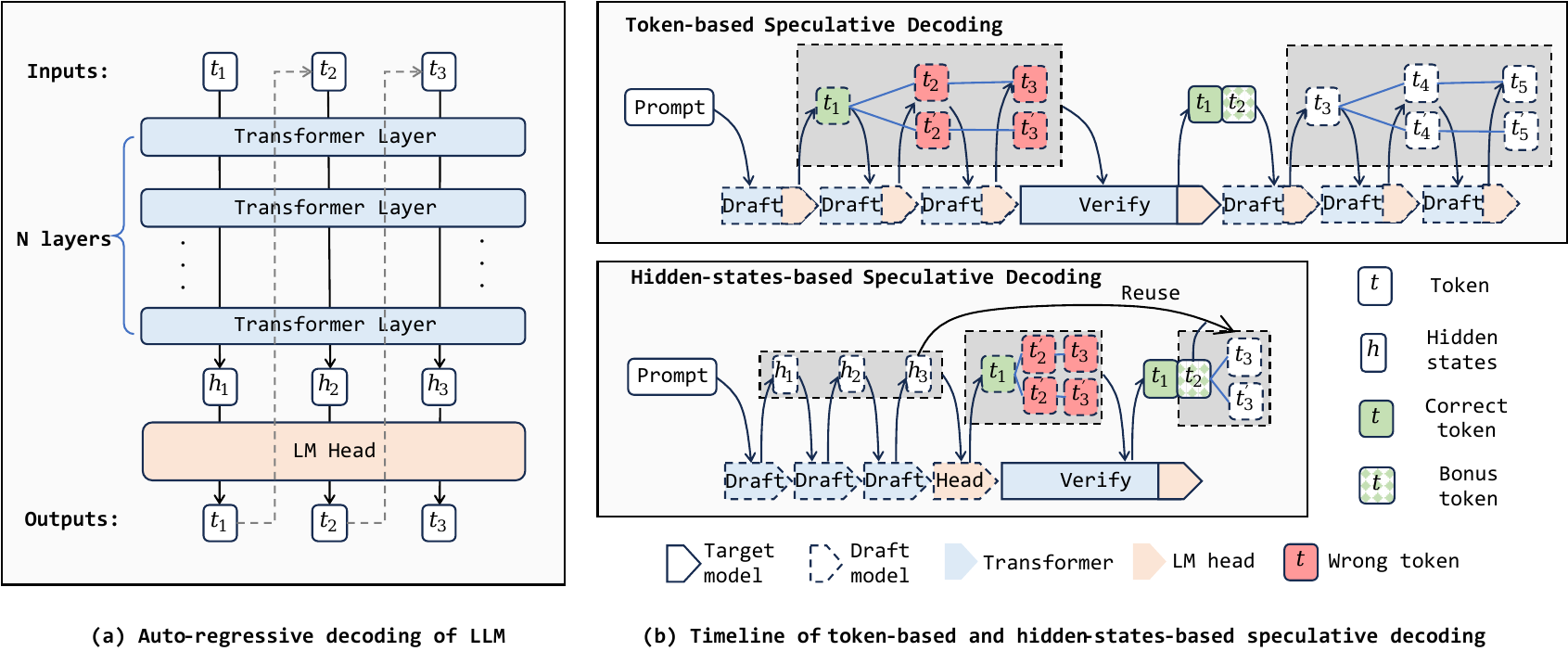}
\end{center}
\vspace{-.5cm}
\caption{Comparing the auto-regressive decoding approach, the token-based speculative decoding approach, and the hidden-states-based speculative decoding approach.}
\label{fig:bg}
\end{figure*}

\subsection{Background}

\textbf{LLM Inference.} As shown in Figure~\ref{fig:bg}(a), transformer-based LLM inference is intrinsically an auto-regressive procedure wherein generating each subsequent token necessitates a sequential model forward pass. This process transforms the inputs through all transformer layers~\cite{10.5555/3295222.3295349}, producing the final hidden states $h$, which are then multiplied by the language model head (LM head) to map them into the vocabulary space as logits. Then, it samples an output token $t$ from the logits. 

The auto-regressive decoding of LLMs results in low arithmetic intensity. Consequently, the throughput of LLM decoding phase is dictated by memory bandwidth rather than compute capability, causing underutilization of GPU computing resources and establishing the critical motivation for optimizations like speculative decoding~\cite{10.5555/3618408.3619203, chen2023acceleratinglargelanguagemodel}.

\textbf{Speculative Decoding.} 
A popular technique to tackle the memory bound is speculative decoding. 
As illustrated in Figure~\ref{fig:bg}(b), we can use a much smaller draft model to speculate, i.e., generate (multiple) candidate tokens~\cite{10.5555/3618408.3619203, chen2023acceleratinglargelanguagemodel}. 
The original model, now serving as the target model, verifies these candidate tokens in parallel, which increases the arithmetic intensity and makes inference compute-bound.

State-of-the-art speculative draft models, represented by EAGLE~\cite{li2024eagle, li2024eagle2}, choose to train specialized draft models to achieve the optimal tradeoff between model size (latency) and accuracy (accepted token length). 
The draft model usually has only one transformer layer to maximally reduce overhead. 
For accuracy, they leverage not only the preceding tokens, but also \textit{hidden states} from the last forward pass of the target model or draft model.
Hidden states represent the model's contextualized understanding of the input that evolves through the layers, containing much richer nuanced information than discrete tokens~\cite{li2024eagle}. 
Thus their integration into the draft models enables EAGLE to achieve longer acceptance lengths; as a result, EAGLE has gained substantial traction in the community~\cite{yan2025scalinglawsspeculativedecoding, tang2025efficientspeculativedecodingllama}.

As shown in Figure~\ref{fig:bg}(b), first proposed by SpecInfer~\cite{10.1145/3620666.3651335}, another widely adopted technique in speculative decoding systems is the tree inference and verification method. Unlike sequential token generation, this approach retains multiple high-probability candidate tokens at each decoding step, organizing them into a tree structure to fully exploit the generated hidden states. The draft tree is then verified in parallel by the target model using a specially designed tree-shaped attention mechanism. By expanding the number of tokens validated in a single forward pass, this tree-based approach significantly increases the expected acceptance length compared to traditional sequential speculative decoding.

\begin{figure}[tb]
\begin{center}
\includegraphics[width=0.7\linewidth ]{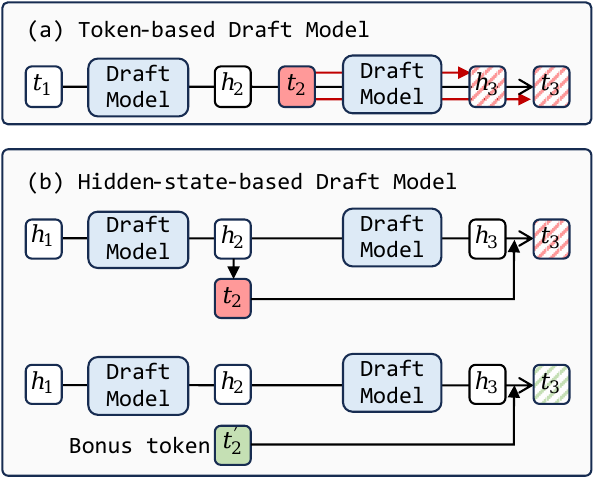}
\end{center}
\vspace{-.5cm}
\caption{For a token-based draft model, an incorrect token implies that all subsequent tokens and hidden states are invalid. In contrast, a hidden-state-based draft model may still retain opportunities for hidden-state reuse.}
\vspace{.5cm} 
\label{fig:insight}
\end{figure}

\subsection{Key Idea}
\label{sec:idea}
Current draft models are auto-regressive with tokens as inputs, following standard language modeling conventions. During inference, they make predictions based on their own previously generated outputs. Consequently, an incorrect token sampling choice invalidates all subsequent hidden states and tokens, since their generation depends on that erroneous token. As illustrated in Figure~\ref{fig:insight}(a), if token $t_2$ fails verification, the subsequent hidden state $h_3$ and token $t_3$ are also incorrect, since $h_3$ was computed conditioned on the rejected $t_2$. Thus, $h_3$ must be discarded and the computation used to generate it is completely wasted.

Our key idea is that it is actually possible to reuse hidden states even when their derived tokens are rejected by the target model once. This can be achieved by decoupling tokens from the draft model's hidden state computation and deferring their incorporation until immediately before sampling, as shown in Figure~\ref{fig:insight}(b). 
This allows us to reuse the same hidden states to generate new tokens without re-executing the full forward pass: when $t_2$ is rejected, $t_3$ is wrong too but $h_2$ and $h_3$ are not affected; we can readily combine the correct token information $t'_2$ from the target model with $h_3$ to sample a new candidate token $t_3'$.
In addition, this eliminates the LM head overhead during the draft model forward phase, which is particularly beneficial for EAGLE-style draft models where the LM head constitutes a significant portion of the total computation.

Intuitively, we believe the new computation paradigm is better suited for draft models: since draft models do not directly output to users, they should focus on auto-regressing their understanding of the context and producing raw hidden states without prematurely collapsing into discrete tokens. Only immediately before final verification do we sample tokens from these hidden states for verification against the target model. \footnote{It remains an interesting open question whether verification can also be based on hidden states directly.}

\subsection{Challenges}

We identify three essential challenges in building a speculative decoding system with hidden-state auto-regressive draft models.

\textbf{Draft model re-design.}
The first challenge is model accuracy. As we remove token information from the draft model's auto-regressive process, its accuracy degrades, affecting the acceptance length and the overall speculative decoding efficiency. 
We empirically demonstrate this degradation in Section~\ref{sec:NHP}.

As shown in Figure~\ref{fig:nhp}, to address this challenge, we train the \textit{next hidden state prediction} model to leverage bonus tokens generated by the target model as initial context. Token information is also injected directly into the sampling process during training, enabling the draft token tree construction and reuse of hidden states via another sampling. 


\textbf{Efficient token information integration.}
To improve the acceptance length, a draft token tree is constructed by sampling with token information.
Then, when a token is rejected, we \textit{re-sample} the logits with the correct token information from the bonus tokens to obtain new candidate tokens to effectively reuse the hidden states.
Thus, our next immediate challenge is how to effectively incorporate the token information because it is the first step for sampling and re-sampling. 
This challenge presents two key difficulties. 
First, the integration of token information during sampling must be extremely efficient, with negligible computational overhead and memory access compared to conventional sampling procedures. 
This is particularly important for re-sampling done over all rejected tokens, for otherwise the re-sampling overhead could outweigh the benefits of hidden state reuse.
Second, since adding token information to logits necessitates a vocabulary-to-vocabulary mapping, a naive approach would require a parameter matrix of size $|V| \times |V|$, where $|V|$ is the vocabulary table, which is prohibitively expensive and unacceptable for practical training. Thus, the number of trainable parameters must be kept small to rein in this cost. 

To address this, we adopt \textit{token-info embedding} (Section~\ref{sec:embedding}). It leverages the canonical low-rank technique to decompose the vocabulary-to-vocabulary mapping into several small linear layers, resulting in a significant reduction in the number of parameters and training time. By composing the low-rank linear layers, we obtain an effective token-info embedding that can be precomputed offline. In the sampling phase, token information is accessed through a simple embedding lookup rather than matrix multiplications, with negligible memory read cost and zero additional computation.

\textbf{System optimization.}  
Finally, we identify the potential performance bottlenecks introduced by our design if not properly addressed.

An acute reader may have already noticed that the token-info embedding relies on a space-time trade-off which may stress the GPU memory. 
To remove any additional computation, token-info embeddings are stored as a matrix of size $|V|^2$. 
For modern LLMs with large vocabularies such as LLaMA-3 with a vocabulary size of 128,256, this amounts to 15.3~GB of HBM usage in FP8. This excessive memory footprint needs to be addressed for practicality.
We exploit the inherent sparsity pattern of the tokens to substantially reduce the memory usage of token-info embeddings as detailed in Section~\ref{sec:sparsity}.

Another notable component is the re-verification needed after re-sampling.
If we simply launch the target model for another round of verification for re-sampled tokens, this overhead could potentially negate the performance gains from hidden state reuse.
Our intuitive solution here is to fuse the re-verification with the ongoing verification of the next batch to effectively amortize the overhead (Section~\ref{sec:fusion}).

\section{Overview of \sys}

We build \sys as a novel speculative decoding system that makes every draft count by efficiently reusing the drafted hidden states. 

\sys employs a hidden state-level auto-regressive model to perform drafting. For each draft iteration, a chain of hidden states is first produced based on target model hidden states and bonus tokens from the last verification. After draft model forward passes, the generated hidden states are projected into logits and token information is injected for sampling, constructing a draft tree with multiple branches. The target model then verifies the draft tree in parallel and outputs bonus tokens. Upon verification completion, \sys identifies reusable hidden states and re-samples new drafts from them, whose verification is merged into the next regular verification.

The rest of the paper is arranged as follows: 
In Section~\ref{sec:DMA}, we describe our draft model based on \textit{next hidden states prediction} which preserves the potential for reusing draft results. In Section~\ref{sec:sampling}, we give a detailed description of how to efficiently injecting token context for prediction and how to construct the draft token tree for sampling and re-sampling. In Section~\ref{sec:overhead}, we introduce \textit{hot-token sparsity} to remove the conflict between HBM footprint and additional computation. Besides, we propose \textit{verification fusion} to further reduce the overhead arising from the verification of re-sampled tokens.

\section{Draft Model Adaptation}\label{sec:DMA}
\begin{figure}[tb]
\begin{center}
\includegraphics[width=\linewidth ]{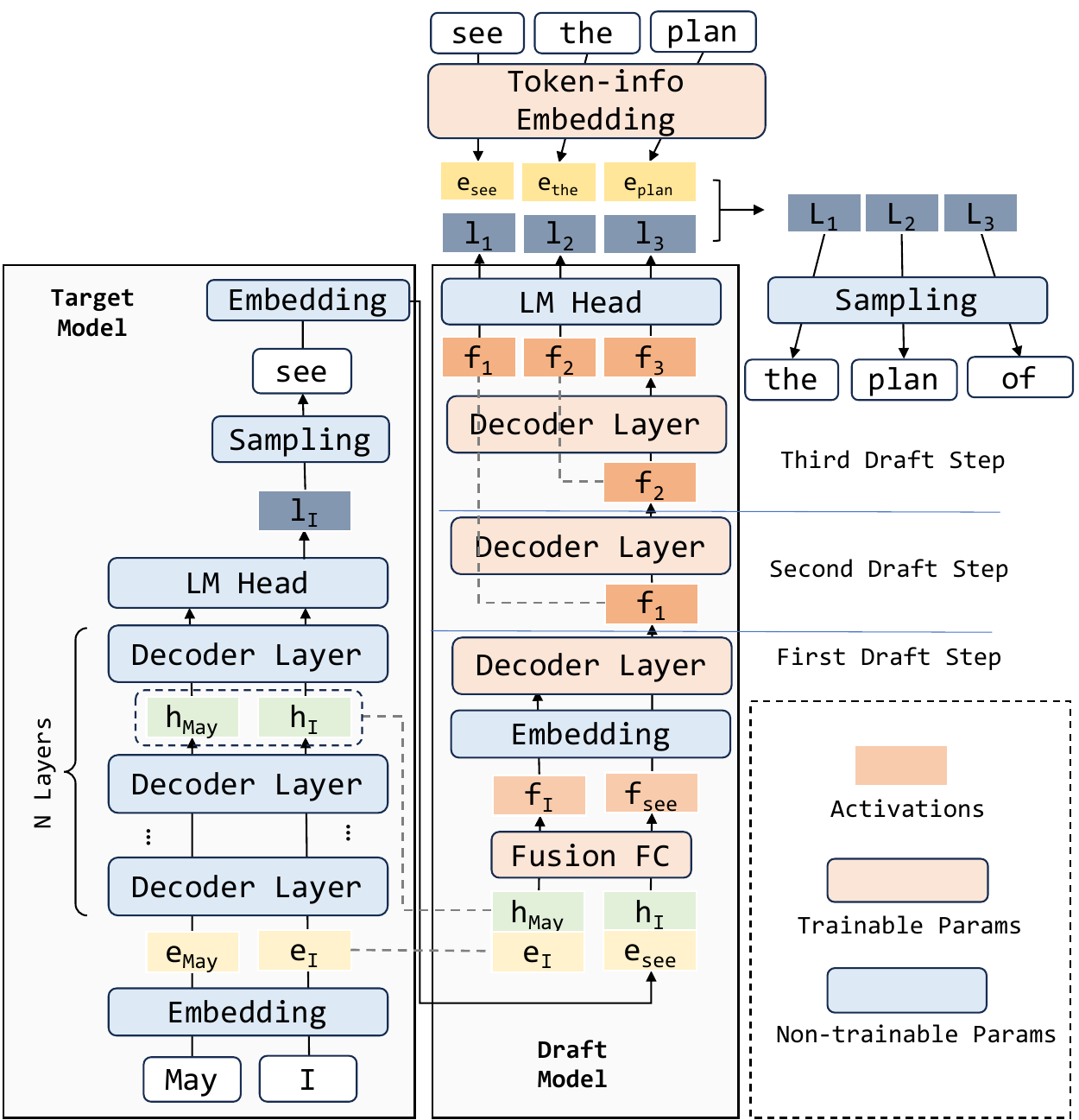}
\end{center}
\vspace{-.5cm}
\caption{Diagram of our draft model inference timeline.}
\label{fig:nhp}
\end{figure}

We have previously established that existing draft models operate as monolithic processes where hidden state updates and token sampling are tightly coupled. The computation of the next hidden state depends strictly on the sampled token. Consequently, if a token choice is rejected during verification, all subsequent computations become invalid. This lack of an intermediate, reusable state prevents the caching of computational trajectories and limits efficiency.

\subsection{Next Hidden State Prediction}
\label{sec:NHP}
To address the limitation, we reshape the drafting process to decouple state evolution from token decoding. By separating the structural generation of the sequence from specific content selection, we create a canonical trajectory of hidden states that remains valid regardless of the specific tokens sampled.

As illustrated in Figure~\ref{fig:nhp}, \sys postpones token conditioning to the final stage of the generation cycle. Specifically, the adapted draft model predicts the next hidden state relying primarily on the current hidden state, effectively eliminating the auto-regressive dependency on sampled tokens during the hidden state generation phase:
\[
h_{i+1} =
\begin{cases}
\text{TransformerLayer}(h_0 \oplus E(t_1)) & i = 0 \\
\text{TransformerLayer}(h_i) & i > 0
\end{cases}
\]
where $E$ denotes the embedding layer of the target LLM, and $\oplus$ represents the fusion operation. Note that only a single transformer layer is deployed with trainable parameters for auto-regression, which is intended for minimizing computation.

\begin{figure}[tb]
\begin{center}
\includegraphics[width=\linewidth ]{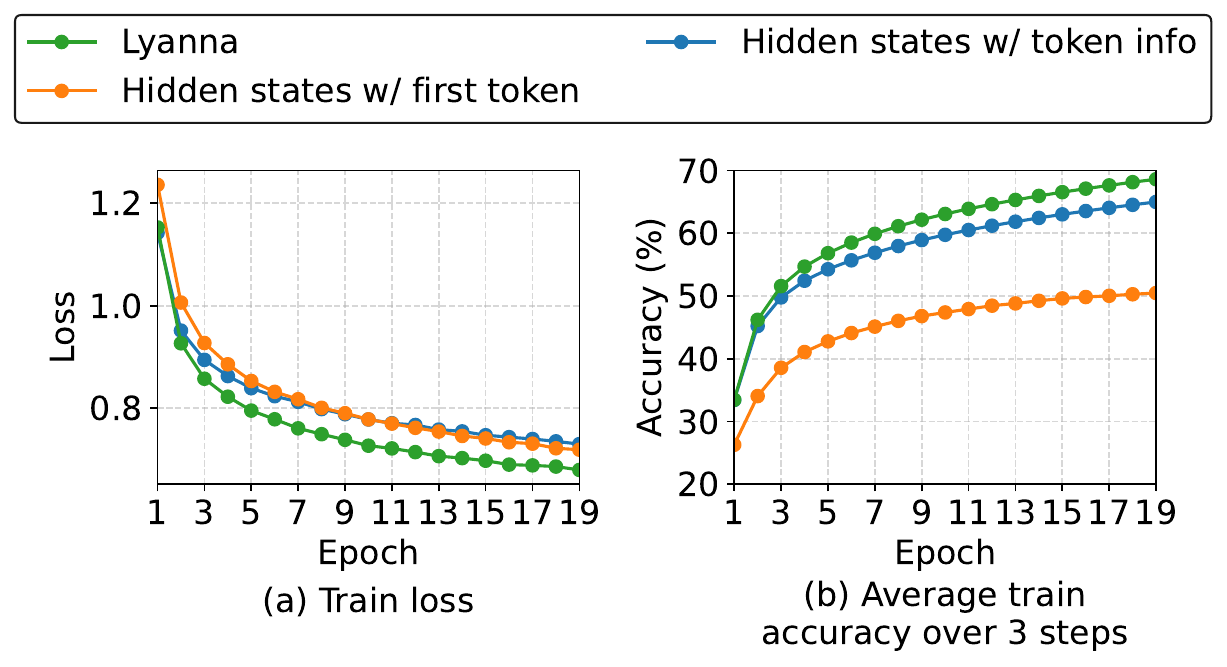}
\end{center}
\vspace{-.5cm}
\caption{Train loss and accuracy across different model optimizations.}
\vspace{.5cm}
\label{fig:trainlossacc}
\end{figure}

An important exception in the draft computation flow occurs at the first draft step ($i=0$), where the draft model initializes its prediction based on the ground truth hidden state and token context provided by the target model. The ground truth token is fused directly into the auto-regression to anchor the sequence. As shown in Figure~\ref{fig:trainlossacc}, integrating this ground truth token is vital for the initial alignment of the draft trajectory.

While the hidden state trajectory ${h_1, \ldots, h_k}$ captures the structural evolution of the sequence, token information remains essential for accurate prediction. To incorporate this without re-introducing auto-regressive dependencies in the state generation, we apply a Normalize-and-Add fusion mechanism at the logit computation stage. We project the hidden states to logits $l_{i+1} \in \mathbb{R}^{|V|}$ (where $|V|$ is the vocabulary size) and fuse the token information as a bias term:
\[
l_{i+1} = h_{i+1}^T \cdot W_{\text{head}}
\]
\[
l'_{i+1} = l_{i+1} + \text{RMSNorm}(E'(t_i))
\]
Here, $E'(\cdot)$ is a separate trainable embedding layer dedicated to this late-stage fusion in the sampling phase, which is detailed in Section~\ref{sec:embedding}.

This architecture yields a significant logical advantage: the hidden state trajectory is canonical. Because the trajectory ${h_1, \ldots, h_k}$ is computed independently of the specific tokens sampled at intermediate steps, the trajectory can be computed once and reused across different sampling outcomes. The additive fusion shifts the output distribution based on the token context, preserving the model's predictive power while enabling the reuse of heavy transformer computations. As demonstrated in Figure~\ref{fig:trainlossacc}, this token conditioning mechanism is critical; we observe a significant improvement in training loss and accuracy compared to hidden-state-only models lacking this fusion.

Training follows a distillation paradigm similar to EAGLE, utilizing a composite loss function that enforces both state regression and distribution alignment:
\[
\mathcal{L} = \alpha \cdot \text{MSE}(h, H) + \beta \cdot \text{CE}(\text{SoftMax}(l'), \text{SoftMax}(L))
\]
where $h$ and $H$ represent the last-layer hidden states of the target and draft models, respectively, and $l'$ and $L$ represent their output logits. We trained the model on 68K samples from the ShareGPT dataset using the Adam optimizer. The training spanned 20 epochs with a learning rate linearly decaying from $3 \times 10^{-5}$ to 0.

\subsection{One-Pass Logits Computation}\label{sec:onepass}
A secondary but critical benefit of decoupling hidden states from token sampling is the parallel computation of the LM head. In standard auto-regressive approaches, the LM head acts as a blocking operation: it must be invoked sequentially because the generation of $h_{i+1}$ depends on the token sampled from $l_i$.

\begin{figure}[tb]
\begin{center}
\includegraphics[width=0.9\linewidth ]{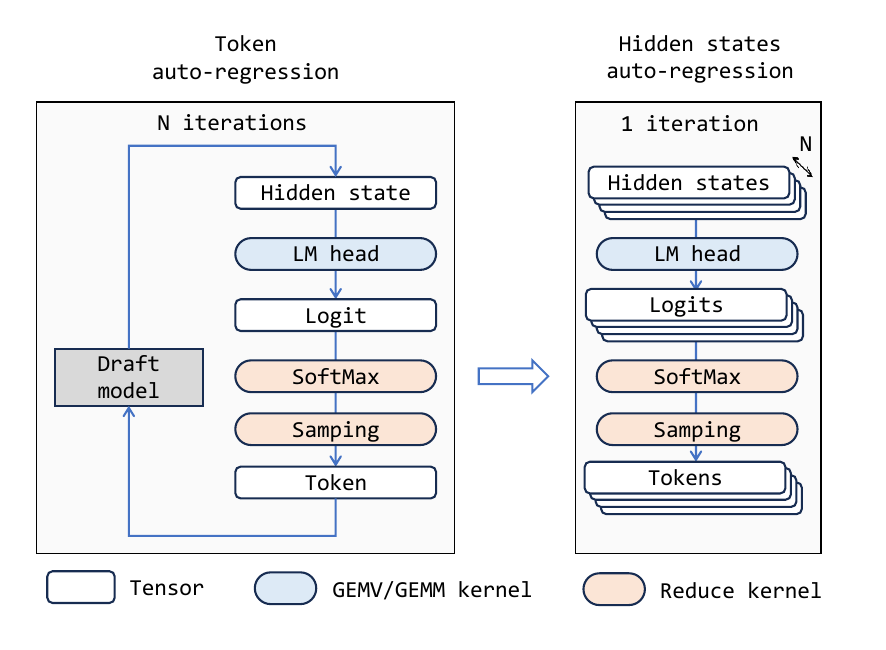}
\end{center}
\vspace{-.5cm}
\caption{Comparison between iterative and one-pass logits computation.}
\vspace{.5cm}
\label{fig:head1}
\end{figure}

However, because \sys computes the hidden state trajectory ${h_1, \ldots, h_k}$ without waiting for intermediate token samples, we can batch these states and propagate them through the LM head in a single pass, as shown in Figure~\ref{fig:head1}:
\[
[l_1, l_2, \ldots, l_k] = [h_1, h_2, \ldots, h_k]^T \cdot W_{\text{head}}
\]
This formulation converts a series of $k$ sequential GEneral Matrix-Vector multiplications (GEMV) into an efficient GEneral Matrix-Matrix multiplication (GEMM). 

\section{Token-info Injected Sampling}\label{sec:sampling}

As mentioned in our key ideas (Section~\ref{sec:idea}), the draft model architecture based on hidden states retains the ability to reuse hidden states by integrating token information. The next essential question is how to realize it with negligible overhead. In this section, we first introduce the token-info embedding which supports sampling with token information without any additional computation. Then, we leverage this technique to construct the draft token tree and reuse hidden states by re-sampling after verification failure.

\begin{figure*}[thp]
\begin{center}
\includegraphics[width=0.9\linewidth ]{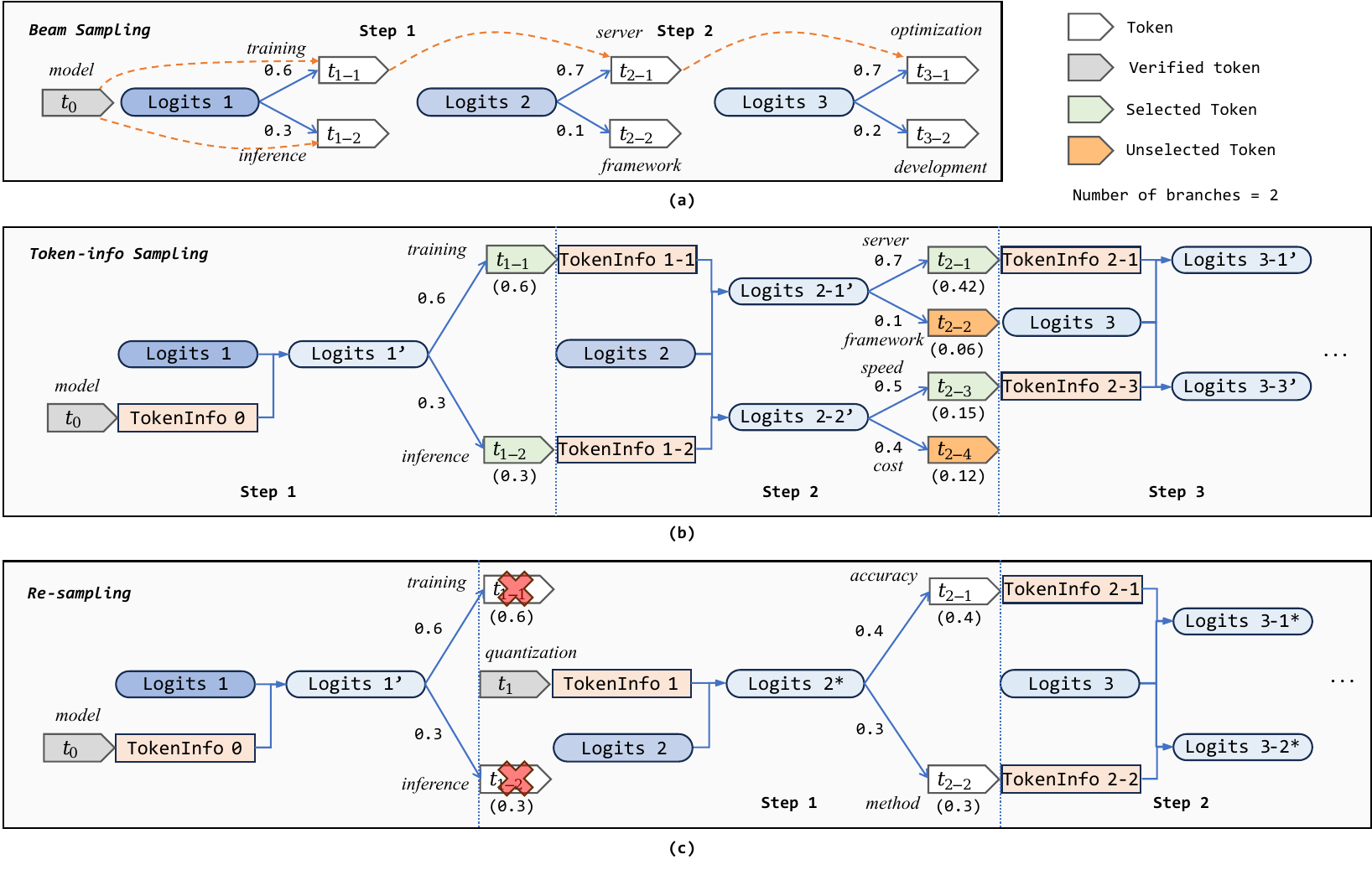}
\end{center}

\vspace{-.5cm}
\caption{Diagram of beam sampling, token-info sampling and re-sampling.}
\vspace{.5cm}
\label{fig:sample}
\end{figure*}

\subsection{Token-info Embedding}\label{sec:embedding}
We have introduced $E'(\cdot)$, termed the \emph{token-info embedding}, to transform discrete token IDs into additive modifications for the logits. To ensure efficient convergence and leverage the semantic knowledge of the target model, we do not train $E'(\cdot)$ from scratch. Instead, we initialize it using a low-rank decomposition. We define two trainable matrices $W_1 \in \mathbb{R}^{n \times d}$ and $W_2 \in \mathbb{R}^{d \times |V|}$, where $d$ is an intermediate dimension and $n$ is the target model's embedding dimension. During training, the token-info embedding is computed as:
\[
E'(t_{i}) = (t_i)^TW_E \cdot W_1 \cdot W_2
\]
where $W_E \in \mathbb{R}^{|V|\times n}$ is the fixed embedding matrix of the target language model.

After training, we collapse this computation by precomputing the product $W_{\text{collapsed}} = W_E \cdot W_1 \cdot W_2$. This materializes the result into a standard lookup table. This design optimizes inference by reducing the complex projection to a simple memory access, eliminating the computational overhead associated with dynamic projections. 

\subsection{Tree Construction by Token-info Sampling }
We need to generate a token tree for verification by sampling from chain-structured logits. Common sampling approaches can sample distinct tokens from multiple logits. However, they cannot construct the draft token tree. The naive approach to address this is to use a variant of beam sampling. This method independently samples tokens from hidden states and assembles the token tree by conducting a cumulative probability-based search. Taking Figure~\ref{fig:sample} (a) as an example and assuming a beam width of 2, we first select the next-layer node \textit{training}, which has the highest probability, starting from the root node \textit{model}. Then, among all visible nodes, the node \textit{server} is selected as it has the maximum probability, calculated as $0.6 \times 0.7 = 0.42$. Subsequently, all tokens become visible. Among them, the node \textit{inference} in the first layer holds the highest probability of 0.3. Following this, the node \textit{optimization} is selected with a cumulative probability of $0.6 \times 0.7 \times 0.7 = 0.294$. This sampling approach proceeds in this manner iteratively to construct the draft token tree.

However, its performance is limited because of the failure to establish causal relationships between tokens across layers. Our proposed token-info sampling addresses this issue by leveraging token information introduced in the model design. As shown in Figure~\ref{fig:sample} (b), the token-info sampling algorithm first derives the token information of \textit{model} and adds it to the first raw logits, which generates the processed logits (\texttt{Logits 1'}). Next, we sample the top-$k$ tokens with the highest probabilities from \texttt{Logits 1'} (where $k=2$ for example), yielding probabilities of 0.6 and 0.3, respectively. Each of these tokens generates its own token information, which is then added to logits 2 to produce distinct logits 2'. We then perform the next round of sampling on these updated logits. Although the top-$k$ tokens are sampled from every distinct logits, the tree does not expand exponentially. Instead, we calculate the cumulative probability for each leaf token. For example, the cumulative probability for the token \textit{server} is $0.6 \times 0.7 = 0.42$. We then select only the two tokens (\textit{server}, \textit{speed}) with the highest cumulative probabilities to continue generating new token information, and this process repeats recursively. The pseudocode for this algorithm is presented in Algorithm~\ref{alg:tree}.

After generating the draft token tree using token-info sampling, \sys prunes the tree based on a verification token budget $B$. If the verification token budget is insufficient to verify the entire draft token tree, the top-$B$ tokens with the highest cumulative probabilities are preserved and forwarded to the target model for verification.

\begin{algorithm}[tb]
\small
\caption{Token-info sampling algorithm used by \sys} \label{alg:tree}
\begin{algorithmic}
\Require Number of speculative steps $N$, hidden states tensor $\mathbf{H}={h_1, h_2, \cdots, h_N}$, draft token tree branch number $k$, last verified token $t_0$, LM head weight $\mathbf{M}$
\Ensure A full token tree $T$
\State \textbf{Structure} \texttt{Node: \{token, prob, jointProb, children\}}
\State $\mathbf{L} \gets \mathbf{H \cdot M}$ \Comment{Convert hidden states to logits}
\State $T \gets \textsc{BuildSubtree}(\mathbf{L}, t_0, k, N)$
\State \Return $T$

\vspace{1em}
\Function{BuildSubtree}{$\mathbf{L}_{\text{start:end}}$, $t_{\text{root}}$, $k$, $N_{\text{steps}}$}
\Require Logits tensor $\mathbf{L}_{\text{start:end}}={l_{\text{start}}, l_{\text{start}+1}, \cdots, l_{\text{end}}}$, root token $t_{\text{root}}$, branch number $k$, number of steps $N_{\text{steps}}$
\Ensure A token subtree rooted at $t_{\text{root}}$
\State $T_{\text{sub}} \gets \texttt{new Node}(t_{\text{root}}, 1, 1, \varnothing)$
\State $Q \gets \texttt{Queue}()$
\State $Q.\texttt{push}(T_{\text{sub}})$

\For{$i=1, 2, \cdots, N_{\text{steps}}$}
    \State $l_i \gets \mathbf{L}_{\text{start}+i-1}$ \Comment{Get logits for current step}
    \State $Q_{\text{next}} \gets \texttt{Queue}()$
    \While{$Q \neq \varnothing$}
        \State $\texttt{node} \gets Q.\texttt{pop}()$
        \State $r_i \gets \texttt{getTokenInfo}(\texttt{node.token})$
        \State $l'_i \gets \texttt{SoftMax}(l_i + r_i)$
        \State $p_1, p_2, \cdots, p_k \gets \texttt{TopkValue}(l'_i)$
        \State $e_1, e_2, \cdots, e_k \gets \texttt{TopkIndex}(l'_i)$
        \State $\texttt{candidates} \gets \varnothing$
        \For{$j=1, 2, \cdots, k$}
            \State $t \gets \texttt{new Node}(e_j, p_j, \texttt{node.jointProb} \times p_j, \varnothing)$
            \State $\texttt{candidates.add}(t)$
            \State $\texttt{node.children.add}(t)$
        \EndFor
        \For{\textbf{each} $t$ \textbf{in} $\texttt{candidates}$}
            \State $Q_{\text{next}}.\texttt{push}(t)$
        \EndFor
    \EndWhile
    \State $Q \gets \texttt{TopkByJointProb}(Q_{\text{next}}, k)$
\EndFor
\State \Return $T_{\text{sub}}$
\EndFunction
\end{algorithmic}
\end{algorithm}

\begin{algorithm}[tb]
\small
\caption{Token-info resampling algorithm} \label{alg:resample}
\begin{algorithmic}
\Require Ground truth token $t_{\text{gt}}$, remaining logits $\mathbf{L}_{\text{remain}}={l_{pos_{\text{reject}}+1}, \cdots, l_N}$, branch number $k$, resample threshold $r$
\Ensure A new token tree $T'$ rooted at $t_{\text{gt}}$
\State \textbf{Structure} \texttt{Node: \{token, prob, jointProb, children\}}

\State $N_{\text{remain}} \gets |\mathbf{L}_{\text{remain}}|$ 
\Comment{Number of remaining steps}

\If{$N_{\text{remain}} > r$}
    \State $T' \gets \textsc{BuildSubtree}(\mathbf{L}_{\text{remain}}, t_{\text{gt}}, k, N_{\text{remain}})$
    \Comment{Build new tree from ground truth token}
\Else
    \State $T' \gets \texttt{new Node}(t_{\text{gt}}, 1, 1, \varnothing)$
\EndIf

\State \Return $T'$ \Comment{Return new resampled tree}
\end{algorithmic}
\end{algorithm}

\subsection{Re-sampling} 
After verifying the draft token tree, a path of the drafted tokens is accepted by the target model. A bonus token is produced by the target model as a byproduct of the verification computation. 

Since identical logits generate identical candidate draft tokens, the processed logits themselves cannot be directly reused. According to the design of token-info sampling, the processed logits used for token generation are derived by adding token information to the raw logits. To leverage these raw logits for improved results, we propose incorporating posterior information obtained from verification into the token information. This updates the processed logits, enabling the reuse of raw logits and the generation of better draft tokens.

Based on the aforementioned observations and insights, we propose \textit{re-sampling}, a novel algorithm designed to reuse draft and verification results to generate higher-quality draft tokens with negligible overhead. 

Figure~\ref{fig:sample} (c) illustrates the re-sampling algorithm. In the scenario where tokens \textit{training} and \textit{inference} are both rejected by the target model during verification, the verification process identifies that the correct token should be the bonus token, \textit{quantization}. The process of standard speculative decoding would end here; \texttt{Logits 2} \& \texttt{3}, along with any subsequent draft outputs, would be discarded. However, the re-sampling algorithm captures the token information of the bonus token (i.e., \textit{quantization} in Figure~\ref{fig:sample}) and sums it with \texttt{Logits 2} to obtain \texttt{Logits 2*}. By incorporating this correct token information verified by the target model, essentially leveraging posterior information, \texttt{Logits 2*} achieves higher quality than \texttt{Logits 2-1'} and \texttt{Logits 2-2'} in Figure~\ref{fig:sample} (b), thereby increasing the probability that its generated tokens will be accepted by the target model. Once \texttt{Logits 2*} is obtained, the subsequent tree construction process proceeds similarly to token-info sampling. Specifically, we select the two tokens with the highest probabilities from \texttt{Logits 2*}, \textit{accuracy} and \textit{method}. We then retrieve their respective token information and add each to \texttt{Logits 3}, yielding \texttt{Logits 3-1*} and \texttt{Logits 3-2*}. This process repeats recursively for subsequent steps. The pseudocode for re-sampling is shown in Algorithm~\ref{alg:resample}.

As detailed in Section~\ref{sec:embedding}, the cost of acquiring token information is minimal; the only additional costs involve vector addition and standard sampling. The overhead of re-sampling is significantly lower than that of constructing a token tree via a new drafting pass. Despite this low overhead, re-sampling proves to be highly effective. Ablation experiments in Section \ref{sec:ablation} show that the re-sampling mechanism brings considerable benefits when the verification overhead is properly handled.

\section{Overhead Removal}\label{sec:overhead}
After introducing draft model adaptation and token-info injected sampling, two overheads hidden in \sys undermine the performance. The first one is the conflict between HBM footprint and computation overhead of token information embedding. The second one is the overhead in verification of resampling results. To remove these two overheads, we introduce \textit{hot-token sparsity} and \textit{verification fusion}.

\subsection{Hot-token Sparsity}\label{sec:sparsity}
In Section~\ref{sec:embedding}, we introduced a low-overhead mechanism for token information retrieval without any computation. This strategy relies on a space-time trade-off between GPU HBM footprint and computation, which requires $O(|V|^2)$ HBM memory.

\begin{figure}[tp]
\begin{center}
\includegraphics[width=0.99\linewidth ]{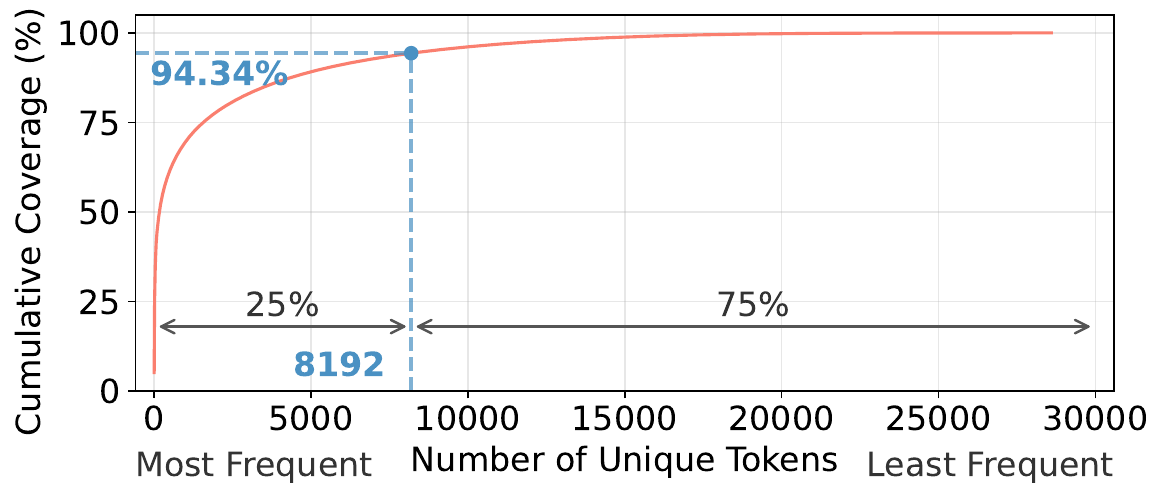}
\end{center}
\vspace{-.5cm}
\caption{Token frequency distribution of LLaMA-2-7B.}
\label{fig:frequency}
\end{figure}

However, some LLMs employ large vocabularies, which makes the HBM requirement for token-info embedding unacceptable. To address this, we observe that the distribution of tokens generated by LLMs is highly skewed. Figure~\ref{fig:frequency} presents the distribution of tokens generated by LLaMA-2-7B on a 1B-token training corpus~\cite{cerebras2023slimpajama}. The results reveal a significant long-tail effect: the vast majority of tokens are rarely used, with 75\% of the vocabulary accounting for only about 5\% of token occurrences. Conversely, a small subset of frequently used tokens, constituting only 25\% of the total vocabulary, accounts for 94.34\% of occurrences. This observation allows us to leverage this sparsity to prune the token information matrix.

We compute token frequency statistics on the training corpus and, for models with a vocabulary size larger than 40K, we retain token-info only for the 32K most frequent tokens when constructing the token-info matrix, in order to control GPU HBM usage. Exploiting this sparsity allows us to prune the token-info matrix to 1/16 of its original size, which removes the conflict between HBM footprint and computation.

\subsection{Verification Fusion}\label{sec:fusion}
The second source of overhead arises in verification. Although the re-sampling algorithm can generate a new candidate token tree with low overhead based on previous hidden states, verifying this tree in the same way as a draft token tree introduces significant overhead. 
This is because the acceptance rate of the re-sampled draft token tree decays more rapidly as the tree depth increases compared to the standard draft token tree. Consequently, verifying the re-sampled token tree should involve fewer tokens because deeper tokens are more likely to be rejected. However, when the batch size and the number of tokens to verify are small, verification becomes memory-bound rather than compute-bound. In this regime, reducing the number of tokens to verify, which decreases computational demand, does not translate to lower verification latency, as memory bandwidth remains the bottleneck.

To resolve this conflict, we propose verification fusion. We merge the re-sampling results from the previous batch into the verification input of the subsequent batch. Since the re-sampling token tree needs relatively few tokens to verify, verification fusion incurs minimal memory bandwidth and computational overhead on the verification process.

\section{Evaluation}

\subsection{Experimental Setup}

\par

\indent \textbf{Implementation}. We implement \sys on top of SGLang~\cite{10.5555/3737916.3739916}, a popular high-performance serving framework for large language models. We use the APIs provided by SGLang to integrate our own draft model, which outputs hidden states instead of token IDs. In addition, we support optimizations such as Paged Attention~\cite{kwon2023efficient}, continuous batching~\cite{280922}, and FlashInfer~\cite{ye2025flashinfer}, matching the configurations used by other baselines.

\textbf{Hardware Platforms}. We have two evaluation platforms. The first platform is equipped with dual Intel Xeon Platinum 8358P processors, 1TB of DDR4 memory, and 4 NVIDIA A800-SXM4-80GB GPUs. The second platform has dual Intel Xeon Platinum 8457C processors, 2TB of DDR4 memory and 4 NVIDIA H800-PCIe-80GB GPUs.

\textbf{Models}. We provide a detailed evaluation of \sys using the LLaMA-2-7B~\cite{touvron2023llama2openfoundation}, one of the most widely-used open-source LLMs. We also demonstrate the applicability of \sys by evaluating it on Vicuna-7B-v1.5 model~\cite{vicuna2023}.
We use BF16 weights and activations for all models~\cite{10.5555/2999134.2999271, 10.5555/3026877.3026899}.

\textbf{Baselines.} We select several state-of-the-art speculative decoding methods supported by modern LLM serving engines as our baselines to achieve their best practical performance, rather than relying on PyTorch- or Transformers-based implementations. The baselines include standard speculative sampling (SPS)~\cite{10.5555/3618408.3619203, chen2023acceleratinglargelanguagemodel, 10.1145/3620666.3651335}, N-gram~\cite{10.1145/3637528.3671614}, and EAGLE~\cite{li2024eagle, li2024eagle2}.

\textit{Standard speculative sampling} employs off-the-shelf models as draft models. Following prior work~\cite{li2025adaserveacceleratingmultislollm, 10.1145/3620666.3651335}, we select LLaMA-160M as the draft model for LLaMA-2-7B, and Tiny-Vicuna-1B as the draft model for Vicuna-7B-v1.5. This approach integrates optimizations from SpecInfer~\cite{10.1145/3620666.3651335}.

\textit{N-gram} leverages a trie-tree–based retrieval mechanism that stores n-gram continuations from prompts and historical generations, enabling extremely fast, context-aware draft construction.
It organizes these drafts into a hierarchical multi-branch token tree and designs a parallel verification procedure.

\textit{EAGLE} is a transformer-based draft model which leverages the output hidden states of the target model. EAGLE features SOTA acceptance length across multiple types of workloads. Following the standard recipe, we train EAGLE2 models for LLaMA-2-7B and Vicuna-7B-v1.5, using the same dataset and training hyperparameters we use to train the \sys draft model.

\textbf{Workload}. We evaluate performance using the benchmark tool provided by SGLang. The evaluation uses a sampling temperature of 0.0 and a top-p value of 1.0. Each run outputs 100,000 tokens across 200 samples. All other settings remain at their default values. We warm up each system with the same configurations. For the N-gram method, since its acceptance length depends on previously completed requests, we warm up the system by choosing requests from the dataset using different random seeds.

\subsection{End-to-end Performance}

\begin{figure*}[t]
\begin{center}
\includegraphics[width=0.9\linewidth ]{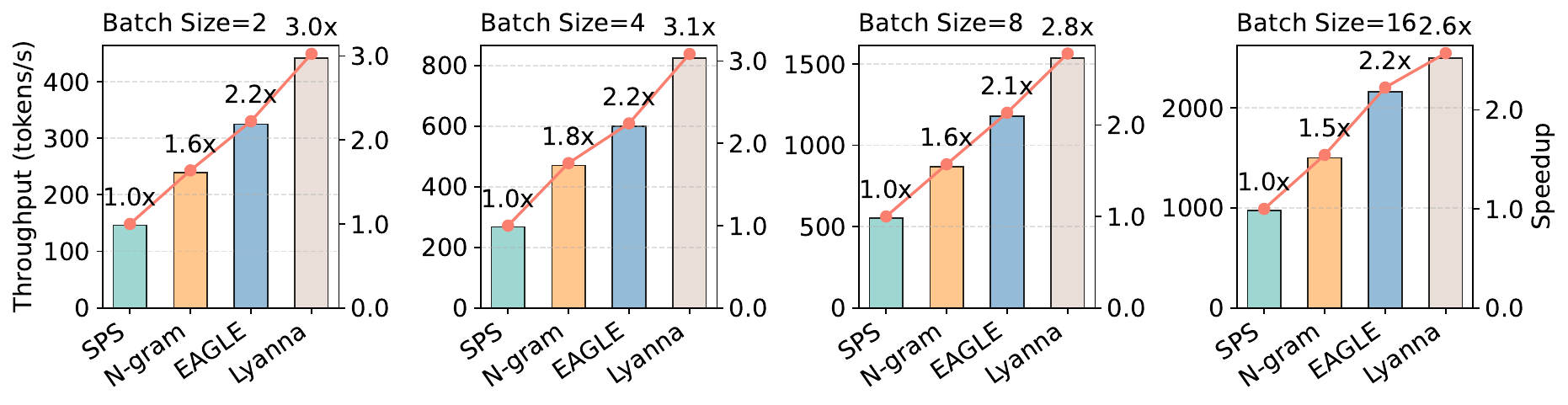}
\end{center}
\vspace{-.5cm}
\caption{Performance comparison of LLaMA-2-7B between state-of-the-arts and \sys on NVIDIA H800 PCIe.}
\label{fig:overall_llama2-7b-h800}
\end{figure*}

\begin{figure*}[t]
\begin{center}
\includegraphics[width=0.9\linewidth ]{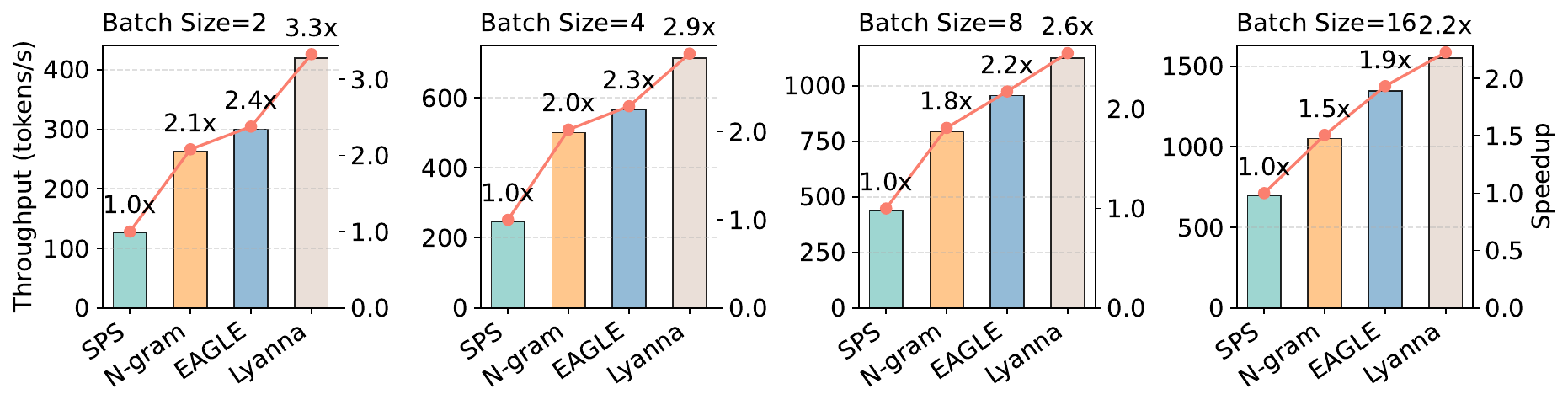}
\end{center}
\vspace{-.5cm}
\caption{Performance comparison of LLaMA-2-7B between state-of-the-arts and \sys on NVIDIA A800 SXM.}
\vspace{.5cm}
\label{fig:overall_llama2-7b-a800}
\end{figure*}

We compare the end-to-end output throughput of LLM inference of different platforms among Standalone, N-gram, EAGLE, and \sys on LLaMA-2-7B and Vicuna-7B-v1.5. The draft token tree has a depth of 3 and a maximum width of 4 nodes per level. Each request verifies 12 tokens per iteration.  For fairness, FlashInfer is employed as the attention backend across all evaluated systems and all other configurations remain the same.

As shown in Figure~\ref{fig:overall_llama2-7b-h800}, \sys delivers the best performance compared to the baselines on LLaMA-2-7B. When batch size is 2, 
Figure~\ref{fig:overall_llama2-7b-h800} and Figure~\ref{fig:overall_llama2-7b-a800} present the output throughput comparison across different batch sizes on both H800 and A800 GPUs. \sys consistently outperforms all baselines across all configurations. On H800 GPUs, \sys achieves throughputs ranging from 442.12 tokens/s at batch size 2 to 2498.64 tokens/s at batch size 16, demonstrating 3.0× to 2.6× speedup over the SPS baseline. Compared to EAGLE, the current state-of-the-art, \sys achieves 1.4× higher throughput at batch size 2, with the performance gap maintained at 1.2× at batch size 16. On A800 GPUs, \sys reaches 1551.88 tokens/s at batch size 16 with speedups of 3.3× to 2.2× over SPS across different batch sizes. Notably, \sys achieves similar performance on H800 and A800 at smaller batch sizes. This is because the NVIDIA H800 PCIe and NVIDIA A800 SXM4 have comparable HBM bandwidth. When batch size is small, speculative decoding throughput is primarily bounded by HBM bandwidth rather than compute capacity. These results demonstrate that \sys achieves superior performance across diverse hardware and batch configurations.

\begin{figure*}[t]
\begin{center}
\includegraphics[width=0.9\linewidth ]{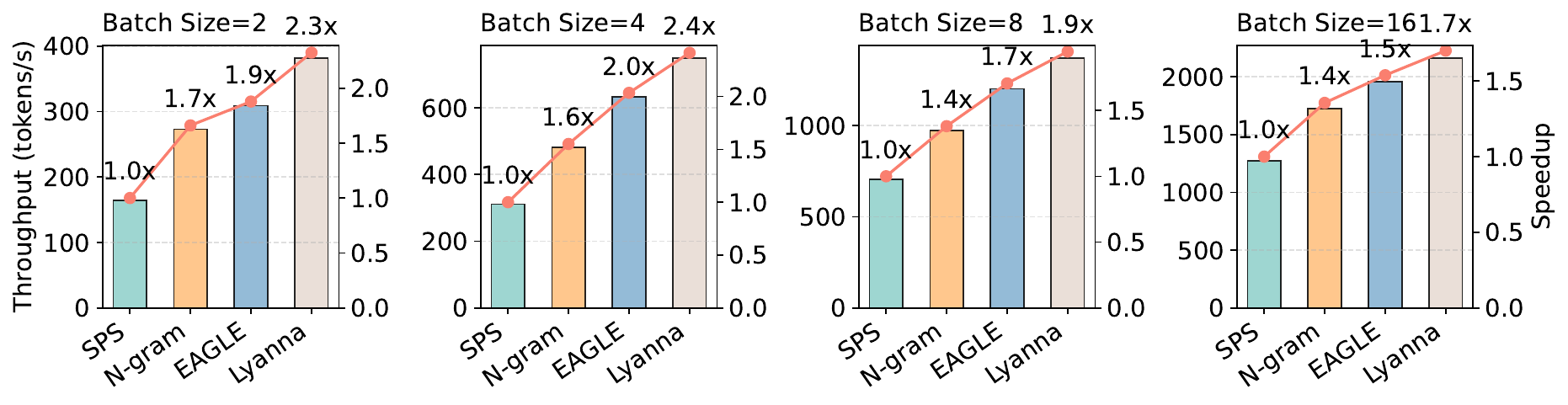}
\end{center}
\vspace{-.5cm}
\caption{Performance comparison of Vicuna-7B-v1.5 between state-of-the-arts and \sys on NVIDIA H800 PCIe.}
\vspace{.5cm}
\label{fig:overall_vicuna-h800}
\end{figure*}

\begin{figure*}[t]
\begin{center}
\includegraphics[width=0.9\linewidth ]{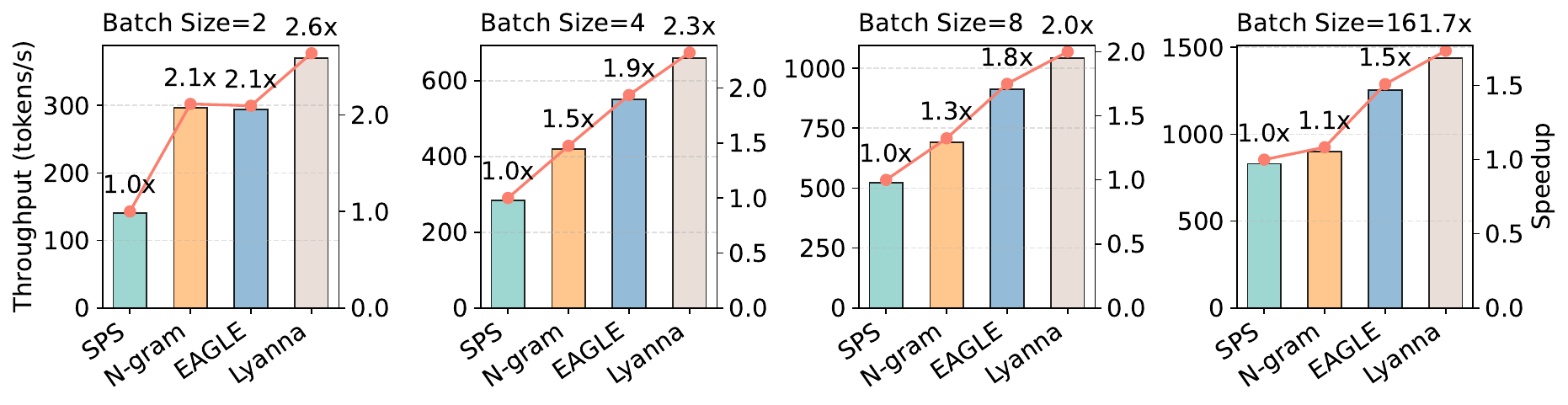}
\end{center}
\vspace{-.5cm}
\caption{Performance comparison of Vicuna-7B-v1.5 between state-of-the-arts and \sys on NVIDIA A800 SXM.}
\label{fig:overall_vicuna-a800}
\end{figure*}

To evaluate the generalizability of \sys across different models, we conduct experiments using Vicuna-7B-v1.5, as shown in Figure~\ref{fig:overall_vicuna-h800} and Figure~\ref{fig:overall_vicuna-a800}. \sys maintains its performance advantage across both hardware configurations. On the H800 GPU, \sys achieves throughputs from 382.17 tokens/s at batch size 2 to 2164 tokens/s at batch size 16, representing 2.3× to 1.7× speedup over SPS. On A800 GPUs, throughput ranges from 370.51 tokens/s to 1439.64 tokens/s with 2.6× to 1.7× speedup. \sys's consistent performance gains across different models demonstrate its robustness and broad applicability to various LLM architectures.

\subsection{Ablation Study}
\label{sec:ablation}
\begin{table}[tb]
\centering
\small
\begin{tabular}{lccc}
\toprule
\textbf{Settings} & \textbf{Step 1} & \textbf{Step 2} & \textbf{Step 3} \\
\midrule
\sys w/o token info & 90\% & 73\% & 64\% \\
\sys w/o first token & 84\% & 71\% & 63\% \\
\sys & 91\% & 80\% & 70\% \\
\bottomrule
\end{tabular}
\caption{Conditional acceptance rate by speculative steps.}
\label{tab:acceptance_rate}
\end{table}

\begin{figure}[tb]
\begin{center}
\includegraphics[width=0.9\linewidth ]{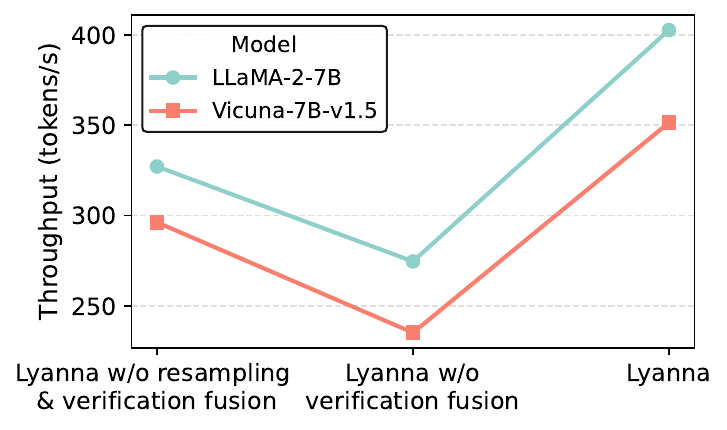}
\end{center}
\vspace{-.5cm}
\caption{Ablation experiments on system optimizations.  } 
\label{fig:ab}
\end{figure}

In this section, we conduct ablation experiments to validate the effectiveness of our design choices for \sys. 

\textbf{Model designs.} We evaluate three variants of our draft model architecture: (1) the complete model with both additive token information and first-step ground truth token fusion, (2) a variant with only token information but without first token, and (3) a variant with only first token fusion but without token information. 

We measure performance using the conditional acceptance rate across the first three speculative steps—that is, conditioned on the previous token being accepted, the probability that the current token is also accepted. The experiment uses a 3-step deep, 4-branch wide tree structure, validating 16 tokens in total. 

As shown in Table~\ref{tab:acceptance_rate}, the complete model with both components consistently achieves the highest acceptance rates across all steps. The results reveal distinct contributions from each design component. The variant with token information demonstrates better stability in maintaining acceptance rates at later steps, with a relatively modest declining trend. In contrast, the variant with only the first token fusion shows a steeper decline, from 90\% to 64\%, despite performing superiorly at step 1. This indicates that while incorporating the first ground truth token provides a strong initial advantage, the additive token information is crucial for sustaining high acceptance rates in deeper speculation steps. The combination of both mechanisms yields the best overall performance, achieving 91\%, 80\%, and 70\% acceptance rates at steps 1, 2, and 3 respectively.

\textbf{System optimizations.} 
Figure~\ref{fig:ab} shows the impact of resampling and verification fusion on throughput for LLaMA-2-7B and Vicuna-7B-v1.5 at batch size 2. Adding resampling alone incurs overhead, reducing throughput by 16.1\% on LLaMA-2-7B and 20.6\% on Vicuna-7B-v1.5 due to the additional verification. However, verification fusion effectively removes this cost. The complete \sys system achieves 402.58 tokens/s on LLaMA-2-7B and 351.37 tokens/s on Vicuna-7B-v1.5, representing 23.1\% and 18.7\% improvements over the configuration without resampling.

\subsection{Draft Model Latency Breakdown}

\begin{figure}[tb]
\begin{center}
\includegraphics[width=0.9\linewidth ]{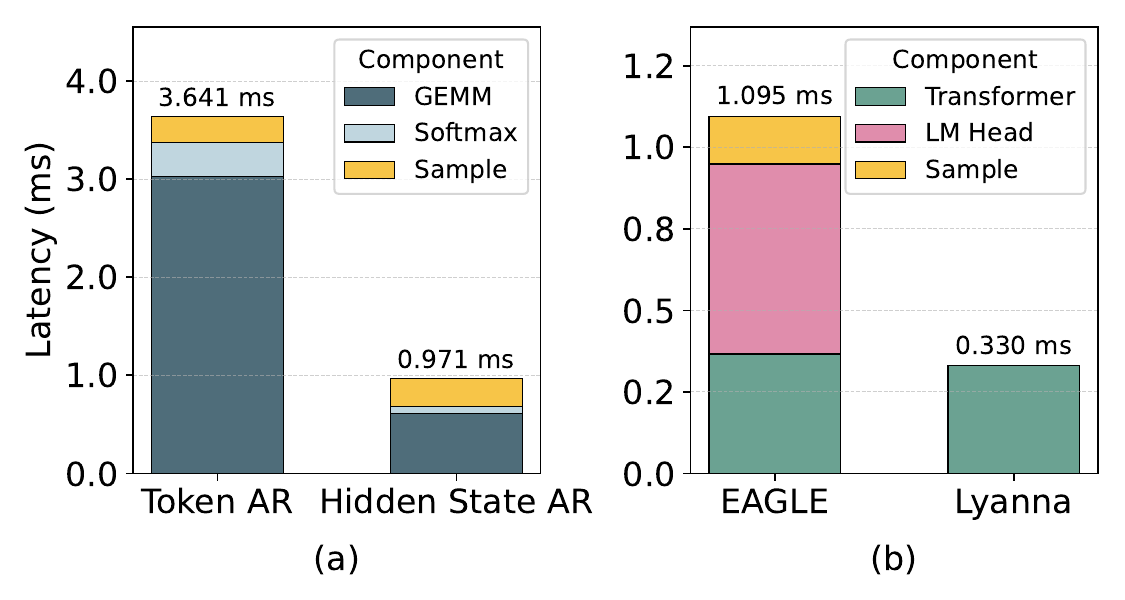}
\end{center}
\vspace{-.5cm}
\caption{(a) Latency comparison between token-based auto-regression (AR) and hidden-state-based auto-regression; (b) draft model forward latency breakdown of EAGLE and \sys.  } 
\vspace{.5cm}
\label{fig:head}
\end{figure}

\textbf{One-Pass Logits Computation Efficiency.}
As described in Section~\ref{sec:onepass}, \sys employs one-pass logits computation by auto-regressively generating hidden states and then performing batched parallel computation of the LM head and SoftMax, thereby increasing arithmetic intensity and reducing latency. Figure~\ref{fig:head} (a) demonstrates this advantage. We measure the latency of LM head computation, SoftMax and sampling at batch size 1 with 5 draft steps. Although batching does not reduce the total amount of computation, the GEMM operations for LM head computation achieve higher arithmetic intensity, resulting in latency that remains nearly constant despite the increased computational workload. The softmax operation exhibits similar behavior. For sampling, \sys's hidden-state-based auto-regression still requires sequential sampling, leading to comparable latency with token-based auto-regressive approaches.

\textbf{Draft Model Forward Latency.}
Figure~\ref{fig:head} demonstrates \sys's efficiency in model forward passes beyond LM head and sampling operations. State-of-the-art methods such as EAGLE minimize draft model forward overhead by reducing the number of transformer layers to just one. However, this design shift makes LM head computation the dominant bottleneck. As shown in Figure~\ref{fig:head}, under SGLang's default configuration, the LM head computation latency exceeds that of the transformer layers. \sys eliminates this overhead entirely by leveraging its hidden-state-based auto-regression, which bypasses repeated LM head computations during draft generation. This architectural advantage makes \sys a significantly more efficient draft model.

\subsection{Hot Token Sparsity}

\begin{table}[tb]
\centering
\begin{tabular}{lcccccccc}
\hline
\textbf{Epoch} & 1 & 2 & 4 & 6 & 8 \\
\hline
\textbf{Sparse} & 42.95 & 50.07 & 56.07 & 60.01 & 60.83 \\
\textbf{Dense} & 42.02 & 48.86 & 55.86 & 58.60 & 59.28 \\
\hline
\end{tabular}
\caption{3-step mean test accuracy comparison between sparse and dense models}
\label{tab:accuracy_comparison}
\end{table}

We evaluate the effect of using only hot tokens in our token-info mechanism. The sparse version restricts token-info embeddings to hot tokens identified through frequency analysis, while the dense baseline uses the full vocabulary.

Table~\ref{tab:accuracy_comparison} shows that the sparse model consistently outperforms the dense counterpart during training, with the performance gap widening from 0.93\% at epoch 1 to 1.55\% at epoch 8. This improvement validates our design choice: focusing on hot tokens not only reduces computational overhead but also enhances generalization by filtering out noise from rare tokens. The results confirm that maintaining token-info only for frequently occurring tokens is both efficient and effective for our speculative decoding framework.

\section{Related Work}

\textbf{Speculative Decoding}. In the area of speculative decoding, there are many system-level optimizations~\cite{10.1145/3695053.3730996, chen2025slosserveoptimizedservingmultislo, 10.1145/3769102.3770608, huang2025specserveefficientsloawarelarge}. Specinfer~\cite{10.1145/3620666.3651335} enables token tree verification. It leverages a tree-based attention masking scheme to fuse parallel verifications of sequences into one CUDA kernel, which greatly improves the efficiency of speculative decoding. AdaServe~\cite{li2025adaserveacceleratingmultislollm} supports multi-SLO LLM serving through SLO-customized speculative decoding. It formulates this scenario as a constrained optimization problem and designs a hardware-aware algorithm to generate a draft token tree which satisfies optimization constraints. Swiftspec~\cite{zhang2025swiftspecultralowlatencyllm} shows that naively reusing the target model’s tensor-parallel strategy for the draft model is inefficient. Instead, SwiftSpec redesigns the speculative decoding pipeline in an asynchronous, decoupled manner so that each component can scale independently, effectively moving draft-model overhead off the critical path.

There are also many algorithmic innovations in the area of speculative decoding~\cite{chen2024sequoia, yi-etal-2024-towards, wu2025stree, gong2025tppsd, sun2024triforce, liu-etal-2024-speculative-decoding, liu2024kangaroo, hu2024accelerated}. 
Moving beyond separate draft models, Medusa~\cite{10.5555/3692070.3692273} introduced a parallel decoding framework that attaches multiple independent heads to the target model's frozen backbone, predicting multiple future tokens simultaneously from the top-layer hidden states. 
{EAGLE}~\cite{li2024eagle, li2024eagle2} and its successors improved upon this by re-introducing auto-regression into the drafting heads; rather than predicting independently, they feed the hidden states of the target model into a lightweight draft layer that accounts for sequential dependencies. 
However, using simplified draft heads introduces distribution shifts. 
{HASS}~\cite{zhang2025learning} addresses the {hidden state exposure bias} caused by shallow draft architectures, while {Griffin}~\cite{hu2025griffin} tackles the complementary {token exposure bias}, ensuring the draft distribution better aligns with the target. 
On the structural efficiency front, {Jakiro}~\cite{huang2025jakiro} replaces dense Feed-Forward Networks with Mixture-of-Experts (MoE) layers in the draft model, demonstrating significant speedups in non-greedy settings. 
Finally, {Cascaded Speculative Drafting}~\cite{10.5555/3737916.3740654} optimizes the verification pipeline by arranging draft models in a vertical cascade, using larger models for immediate tokens and progressively smaller models for deeper positions.

\section{Conclusion}
In this paper, we present \sys, a novel speculative decoding system that enables reuse of computation from rejected tokens. \sys tackles the key challenge of computational waste in speculative decoding by introducing a hidden state chain mechanism and token-info injected sampling. Our approach achieves up to 3.3× speedup over standard auto-regressive decoding and outperforms existing speculative methods while maintaining output quality. Extensive experiments across multiple benchmarks demonstrate that \sys offers a practical and efficient solution for accelerating large language model inference.

\bibliographystyle{plain}
\bibliography{osdi}

\end{document}